\documentclass{article}

\usepackage{arxiv}

\usepackage[utf8]{inputenc} 
\usepackage[T1]{fontenc}    
\usepackage{hyperref}       
\usepackage{url}            
\usepackage{booktabs}       
\usepackage{amsfonts}       
\usepackage{nicefrac}       
\usepackage{microtype}      
\usepackage{lipsum}
\usepackage{graphicx}
\usepackage{caption}
\usepackage{subcaption}
\usepackage{float}
\usepackage{amsthm}
\usepackage{enumitem}
\graphicspath{ {./images/} }

\usepackage{xcolor}
\usepackage[noend,ruled,algonl]{algorithm2e}
\usepackage{amsmath}
\usepackage{nicematrix}
\usepackage{tkz-graph}
\usepackage{bm}
\usepackage{tikz}
\usepackage{cancel}
\usepackage{algpseudocode}
\usepackage{listings}

\lstdefinestyle{cStyle}{
    language=C,
    basicstyle=\ttfamily\small,
    keywordstyle=\color{blue},
    commentstyle=\color{gray},
    stringstyle=\color{teal},
    numbers=none,
    frame=none,
    breaklines=true,
    showstringspaces=false,
    tabsize=4,
    columns=fullflexible
}

\lstdefinestyle{pythonStyle}{
    language=Python,
    basicstyle=\ttfamily\small,
    keywordstyle=\color{blue},
    keywordstyle=[2]\color{violet},
    keywordstyle=[3]\color{orange},
    stringstyle=\color{teal},
    commentstyle=\color{gray},
    numbers=none,
    frame=none,
    showstringspaces=false,
    breaklines=true,
    columns=fullflexible,
    morekeywords=[2]{SSAKG},
    morekeywords=[3]{True,False,None}
}

\usetikzlibrary{decorations.pathreplacing}

\definecolor{purple_1}{HTML}{E4CCFF}
\definecolor{green_1}{HTML}{339955}
\definecolor{green_2}{HTML}{EBFFF2}
\definecolor{orange_1}{HTML}{FEA333}

\newcommand{\CFunction}[1]{{\textcolor{green_1}{#1}}}

\SetKwProg{Prioritize}{\CFunction{\textbf{prioritize}}}{}{end}
\SetKwProg{OrderingAlg}{node\_ordering}{}{end}
\SetKwProg{GetUnorderedSequence}{get\_unordered\_sequence}{}{end}
\SetKwProg{FilterSequenceMatrix}{filter\_sequence\_matrix}{}{end}
\SetKwProg{GetOrderedSequence}{get\_ordered\_sequence}{}{end}

\newcommand{\myapp}{\texttt{SSAKG 2.0 }}

\title{\myapp: An Open-Source Package for Structural Associative Sequence Memory and Context-Based Retrieval}

\author{
    Przemysław Stokłosa \\
    Institute of Management and Information Technology\\
    Bielsko-Biała, Poland \\
    \texttt{przemyslaw.stoklosa@gmail.com} \\
    \And
    Janusz A. Starzyk \\
    University of Information Technology and Management\\
    Rzeszów, Poland\\
    \texttt{starzykj@gmail.com} \\
    \And
    Paweł Raif \\
    Silesian University of Technology\\
    Gliwice, Poland\\
    \texttt{pawel.raif@gmail.com} \\
}

\newtheorem{example}{Example}

\begin{document}
    \maketitle
    \begin{abstract}

This article presents \texttt{SSAKG 2.0}, an open-source software package for constructing and operating Structural Sequential
Associative Knowledge Graphs ($SSAKGs$).
An $SSAKG$ represents objects as graph
vertices and ordered sequences as structural patterns of graph connections.
The resulting sparse graph is used as an associative memory in which complete sequences can be
reconstructed from a partial, unordered context.
Version 2.0 introduces new algorithms that exploit individual bits of computer memory to
efficiently search graph connections.
The package is implemented in $Python$, while performance-critical graph
operations are implemented in $C$ and exposed through a $Python$ interface.
This hybrid implementation provides a flexible high-level programming environment
while reducing the memory and computational overhead associated with large sparse graphs.
The algorithms were evaluated using randomly generated numerical sequences, sequences derived from
sentences in the \textit{NLTK }corpus and \textit{mRNA} sequences.
The experiments demonstrate the ability of the package to store and reconstruct
sequences from partial contexts and provide a basis for evaluating the effects
of graph density, sequence length, and memory size on retrieval performance.
\myapp is distributed under the Apache 2.0 open-source license.
The package includes documentation and reproducible examples and is
publicly available through GitHub and the Python Package Index (PyPI).
    \end{abstract}
    \keywords{sequence retrieval \and context-based association \and associative knowledge graphs \and graph density \and ssakg package}

    \section{Introduction}

    Associative memory provides a mechanism for retrieving stored information from partial,
    incomplete, or noisy cues rather than from an explicit address.
    This idea has played an important role in neural computation and cognitive modeling,
    leading to a variety of architectures for content-addressable memory, including early correlation-based associative
    memories, Hopfield networks and sparse distributed memory
    ~\cite{hopfield1982neural,kanerva1988sparse,keeler1988comparison,liang2022modern,wu2024uniform}.
    Sparse Distributed Memory is particularly
    relevant to the present work because it demonstrates how sparse connectivity
    can be used to obtain associative recall while maintaining a relationship
    between storage capacity and the number of memory connections
    ~\cite{willshaw1969nonholographic,keeler1988comparison}.
    Although these approaches differ substantially
    in their representations and retrieval mechanisms, they share the general
    objective of recovering stored information on the basis of its content or partial contextual information.
    A related but distinct problem is the associative storage and retrieval of ordered sequences
    ~\cite{Kak1990SelfIndexing, Zhang2023BIAA,Lan2021SpatioTemporal, abolpour2021}.
    In many applications, the information to be recalled
    is not a single static pattern but an ordered collection of elements whose temporal or positional relationships are essential.
    Examples include symbolic and numerical sequences, natural-language expressions, biological sequences, behavioral
    trajectories, and temporal event streams.
    In such cases, an associative memory must address two related tasks: identifying the elements associated with a partial
    context and reconstructing, when required, their original order.
    Sequence storage and retrieval have therefore been investigated using a variety of
    recurrent neural networks, associative memories, and external-memory architectures.

    Structural Associative Knowledge Graphs (SSAKGs)
    address this problem using a graph-based representation in which objects are
    represented by vertices and sequences are encoded through structural
    relationships between them.. Rather than storing every sequence as an
    independent record, multiple sequences are superimposed within a common graph and may share vertices and connections.
    A stored sequence can therefore be represented as a
    structural pattern embedded in the graph, while retrieval is initiated from a context consisting of a subset of its elements.
    The context may be incomplete and does not necessarily have to preserve the original order of the elements.

    The SSAKG model and its basic sequence-reconstruction algorithms were introduced in our previous work
    ~\cite{stoklosa_ssakg}.
    That study investigated the relationship between
    graph density, memory capacity, and retrieval accuracy and demonstrated that
    structural graph representations can support context-based sequence
    reconstruction without storing each sequence as a separate record.
    Subsequent work further ~\cite{raif2026} investigated the scalability of sparse
    graph-based associative memory and compared this approach with modern
    Hopfield-type architectures~\cite{ramsauer2020hopfield}.
    We demonstrate that our approach significantly improves associative memory capacity and search accuracy compared to MHN,
    while requiring less computer storage for very large memories.

    The objective of the present work is different.
    Rather than introducing a new associative-memory model, this article presents \myapp,
    an open-source and reusable software implementation of the \texttt{SSAKG} framework
    and its associated algorithms.
    The package provides a practical environment for constructing, modifying,
    and querying structural associative memories and for
    experimenting with alternative sequence representations and retrieval procedures.
    In this sense, the principal contribution of the present work is
    not only an algorithmic description, but the transformation of the previously
    developed model into a documented, extensible, and reproducible software framework.

    A particular focus of \myapp is the efficient representation and processing of sparse graph connectivity.
    Because graph connections are represented by binary information, individual bits of computer
    memory can be used to encode and search multiple structural relationships efficiently.
    The package combines a high-level Python interface with
    performance-critical operations implemented in \texttt{C}.
    This design allows users to construct and manipulate associative memories through a relatively simple
    programming interface while performing selected graph operations at the bit level.

    The resulting approach differs from architectures in which memory is implemented as an external
    differentiable resource accessed through an addressing mechanism, such as Neural Turing Machine
    ~\cite{graves2014neural} and Differentiable Neural Computer~\cite{graves2016hybrid,csordas2019improving}
    serves as a good point of contrast: memory is an external
    resource accessed by the network via an addressing mechanism, whereas In SSAKG,
    the graph itself constitutes the associative memory: the stored information is
    represented directly by the connectivity structure, and retrieval is performed
    by exploiting structural relationships between contextual elements and candidate sequence components.

    The package is intended as a general research and experimentation tool rather than as a domain-specific application.
    The stored elements may represent arbitrary symbols, numerical identifiers, words, or other discrete objects.
    This allows the same underlying implementation to be
    applied to different classes of sequence data.
    In the present study, SSAKG is evaluated primarily through controlled experiments designed to examine the
    effects of memory size, graph density, sequence length, and context size on retrieval performance.
    We have provided additional examples on SSAKG’s GitHub page~\cite{Sto2024} to illustrate how the
    same representation can be applied to a variety of symbolic data using
    linguistic and biological sequences.

    The main contributions of this work are as follows:
    \begin{enumerate}[label=(\roman*), leftmargin=*]
        \item{
            \textbf{An open-source implementation of Structural Associative Knowledge Graph memory}
            for storing and retrieving ordered sequences from partial contextual information.
        }
        \item{
            \textbf{A reusable software architecture} that separates the high-level Python
            interface from performance-critical low-level graph operations.
        }
        \item{
            \textbf{Bit-level algorithms for representing and searching graph connectivity}
            , implemented through a $C$ extension and integrated with the $Python$ package.
        }
        \item{
            \textbf{Algorithms supporting context-based retrieval and ordered sequence reconstruction},
            including the handling of repeated sequence elements.
        }
        \item{
            \textbf{An experimental evaluation of the package}
            and its retrieval algorithms under varying memory and sequence conditions.
        }
        \item {
            \textbf{A reproducible and extensible software distribution},
            including documentation, examples, source code, and package availability
            through public repositories.
        }
    \end{enumerate}

    The remainder of the paper is organized as follows.
    Section~\ref{sec:preliminaries} the basic concepts and notation underlying SSAKG memory.
    Sections~3--7 describe the sequence representation, storage procedures, retrieval algorithms, and software implementation.
    Section~\ref{sec:experimental-evaluation} presents the experimental evaluation.
    Section~\ref{sec:discussion} discusses the computational properties and limitations and possible future extensions of the package.
    Finally, Section~\ref{sec:conclusions} summarizes the main conclusions.

    \section{Preliminaries}
    \label{sec:preliminaries}
    Classical associative memories often
    focus on pattern retrieval, whereas SSAKG is designed for the retrieval of ordered sequences.
    The problem addressed by SSAKG is not merely pattern
    completion, but the reconstruction of an ordered sequence from a partial and
    potentially unordered subset of its elements.
    Rather than storing sequences as separate records, SSAKG encodes them in the topology of a shared graph.
    Several sequences may share common elements, so there is no need to store each sequence independently.
    Shared objects become shared nodes in a graph, and the
    connection structure encodes the relationships arising from the occurrence of
    objects within the sequences~\cite{stoklosa_ssakg}.

    \subsection{Symbols, sequences and contexts}
    \label{subsec:symbols-sequencesand-contexts}

    The associative memory $SAM$ enables the storage of ordered strings of symbols, called $S$ sequences.
    A sequence can be retrieved by providing its fragment.
    This fragment may be unordered.
    Such an unordered fragment is called context $C$. \\
    Initially, we select a set of symbols $N$.
    The SSAKG memory stores ordered sequences of symbolic objects.
    Let
    \begin{equation*}
        N = \{n_1,n_2,\dots,n_k\}
        \label{eq:equation_symbols}
    \end{equation*}
    denote the set of available symbols.
    A sequence $S$ is an ordered tuple:
    \begin{equation*}
        S = (s \in N)
        \label{eq:equation_sequence}
    \end{equation*}
    In order to read a sequence from memory, we need to specify a fragment of it, so-called context:
    \begin{equation*}
        C = \{c \in S\}
    \end{equation*}
    for example, a sequence of length 5 can take the form:
    \begin{equation*}
        S = (n_2,n_1,n_3,n_{10},n_1)
    \end{equation*}
    while the context required to read this sequence may be a set of symbols:
    \begin{equation*}
        C = \{n_{10},n_1,n_3\}
    \end{equation*}

    \subsection{Graph representation of sequences}\label{subsec:graph-representation-of-sequences}
    The associative memory is represented by a graph $G_{sam}=(V,E)$ with $k=|N|$ vertices.
    Each vertex $v\in V$ is associated with the corresponding symbol $n\in N$.
    This graph forms the basis of the associative memory.
    Its adjacency matrix $A_{sam}$ is stored in the computer’s memory.

    When a sequence is stored, its elements are mapped to
    vertices and the relationships required to represent their order are added to
    the associative graph.

    Let $G_s$ denote the graph structure associated with sequence $S$,
    and let $G_{sam}$ denote the graph representing the complete associative memory.
    Storing $S$ consists of incorporating the structural relationships of $G_s$ into $G_{sam}$:
    \begin{equation*}
        G_{sam}\leftarrow G_{sam}\cup G_s
    \end{equation*}
    The exact structure of $G_s$ determines the information available during retrieval and
    therefore has a direct influence on the efficiency, robustness, and capacity of
    the associative memory.
    The graph can be represented by an adjacency matrix $A_{sam}$.
    Initially, the graph contains no edges; the matrix is filled with zeros.
    The appropriate edges are created when a sequence is stored.
    A sequence is represented by the corresponding graph $G_s$.
    The elements of the sequence are the vertices of the graph, arranged in a given order.
    The key element is the choice of the graph $G_s$ representing the sequence.
    It is the structure of this graph that enables the construction
    of fast and efficient algorithms for reading the sequences.
    The appropriate vertices, together with the edges of the graph $G_s$, are inserted into the graph $G_{sam}$.
    The $A_{sam}$ matrix has been modified as follows:
    \begin{equation}
        A_{sam}[\{v_1,v_2,\cdots,v_i\}]=A_s
        \label{eq:submatrix_asam}
    \end{equation}
    where $A_{sam}[\{v_1,v_2,\cdots,v_i\}]$ denotes the submatrix of the matrix $A_{sam}$ corresponding to
    the set of vertices $\{v_1,v_2,\cdots,v_i\}$.
    The simplest case considered is the use of a complete graph.
    This is described in more detail in ~\cite{Starzyk2024_}.
    If the sequence contains repeating symbols, new vertices are created in the graph for them.
    The matrix $A_{sam}$ is dynamically expanded.

    \section{Sequences storage}\label{sec:sequences-storage}

    \subsection{Motivation for \myapp}\label{subsec:motivation-for-ssakg-1.0}
    The previous work established the
    SSAKG model and evaluated its sequence reconstruction algorithms~\cite{stoklosa_ssakg, raif2026}.
    The objective of \myapp is different: to provide an efficient, reusable, and extensible
    software implementation of the model and its associated algorithms.

    The SSAKG package implements an associative memory based on a sparse graph representation.
    Sequences are stored by establishing structural relationships between the graph vertices
    corresponding to their elements.
    The resulting graph is shared by all stored
    sequences, allowing sequences to be retrieved from partial contextual information.
    An important feature of the implementation is its ability to
    preserve the order of sequence elements while also supporting sequences
    containing repeated symbols.

    The following example illustrates the basic principle of sequence storage implemented in the package.
    It also provides a simple representation of the memory structure used by the retrieval
    algorithms described in the subsequent sections.

    \subsection{Basic sequence representation}
    \label{sec:zapis-sekwencji}

    Consider an associative memory graph $G_{sam}$ containing $6$ vertices $\{1, 2, 3, 4, 5, 6\}$ and initially no edges.
    Its adjacency matrix is therefore a $6 \times 6$ matrix containing zeros.
    A sequence containing three elements can be represented by a complete graph $G_s$ with three vertices, whose adjacency
    matrix is $A_s = K_3$.

    It should be noted that the graph $A_s$ establishes
    associations between all elements of the sequence.
    However, this representation does not encode their order.
    Consequently, although the set of symbols associated with a stored sequence can subsequently be identified, their
    original ordering cannot be recovered from this representation alone.

\begin{figure}[H]
        \centering
        \begin{subfigure}[t]{0.45\textwidth}
            \centering
            \caption{Sequence graph}
            \begin{tikzpicture}[scale=0.6]
                    \GraphInit[vstyle=Normal]
                    \SetGraphUnit{1}
                    \begin{scope}
                        \Vertices{circle}{1,2,3}
                    \end{scope}
                    \Edges(1,2,3,1)
            \end{tikzpicture}
        \end{subfigure}
        \begin{subfigure}[t]{0.45\textwidth}
            \centering
            \caption{Adjacency matrix}
            \[
                 A_s=
                 \begin{pmatrix}
                     0 & 1 & 1\\
                     1 & 0 & 1\\
                     1 & 1 & 0\\
                 \end{pmatrix}
            \]
        \end{subfigure}
        \caption{A graph representing a sequence, together with its adjacency matrix}
        \label{fig:sequence_graph_1}
    \end{figure}
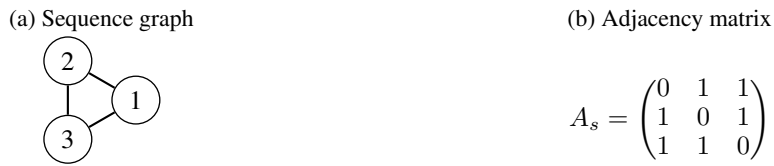

    The graph $G_s$ and the matrix $A_s$ are shown in Figure~\ref{fig:sequence_graph_1}.

    \begin{example}
        \label{example:1}

        As an example, let us take the two sequences $\{2,6,1\}$ and $\{6,4,5\}$.
        A memory constructed in this way allows the set of symbols in the sequence to be read, but the order is random.

        The graph $G_s$ with three vertices will be embedded in the graph $G_{sam}$ at two example positions.
        The matrix $A_s$ will be inserted into the submatrix of the graph $A_{sam}$ corresponding
        to the sets of vertices $\{2,6,1\}$ and $\{6,4,5\}$.
        In matrix notation, this can be written as:
        \[
    A_{sam} [\{v_2,v_6,v_1\}] = \begin{pmatrix}
                                    0 & 1 & 1\\
                                    1 & 0 & 1\\
                                    1 & 1 & 0\\
    \end{pmatrix}=A_s, A_{sam} [\{v_6,v_4,v_5\}]=A_s
\]
        where $\{v_2, v_6, v_1\}$ and $\{v_6, v_4, v_5\}$ correspond to the vertices where the graphs $G_s$ representing the respective sequences will be placed.
        The graph $G_{sam}$, with its adjacency matrix $A_{sam}$, is shown in Figure~\ref{fig:sequence_graph_2}
        \NiceMatrixOptions{
    code-for-first-row = \color{red},
    code-for-first-col = \color{blue},
    code-for-last-row = \color{green},
    code-for-last-col = \color{magenta}
}

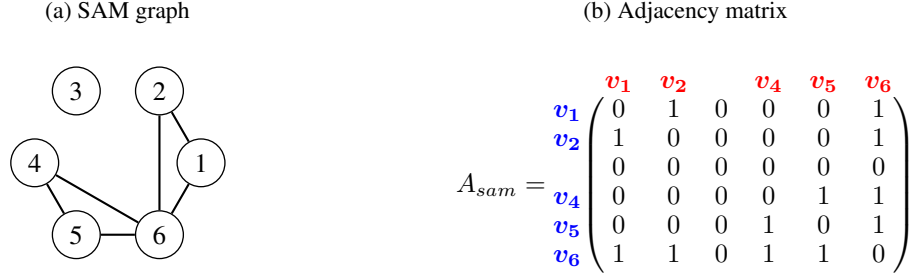
\begin{figure}[H]
    \centering
    \begin{subfigure}[t]{0.45\textwidth}
        \centering
        \caption{SAM graph}
        \vspace{0.5cm}
        \begin{tikzpicture}[scale=1.1]
            \GraphInit[vstyle=Normal]
            \SetGraphUnit{1}
            \begin{scope}
                \Vertices{circle}{1,2,3,4,5,6}
            \end{scope}
            \Edges(1,2,6,1)
            \Edges(6,4,5,6)
        \end{tikzpicture}
    \end{subfigure}
    \begin{subfigure}[t]{0.45\textwidth}
        \centering
        \caption{Adjacency matrix}
        \[
            A_{sam}=
            \begin{pNiceMatrix}[first-row,first-col]
                & \bm{v_1} & \bm{v_2} &  & \bm{v_4} & \bm{v_5} & \bm{v_6} \\
                \bm{v_1} & 0 & 1 & 0 & 0 & 0 & 1 \\
                \bm{v_2} & 1 & 0 & 0 & 0 & 0 & 1 \\
                & 0 & 0 & 0 & 0 & 0 & 0 \\
                \bm{v_4} & 0 & 0 & 0 & 0 & 1 & 1 \\
                \bm{v_5} & 0 & 0 & 0 & 1 & 0 & 1 \\
                \bm{v_6} & 1 & 1 & 0 & 1 & 1 & 0
            \end{pNiceMatrix}
        \]
        \label{fig:sequence_graph_2:adjacency}
    \end{subfigure}
    \caption{The graph $G_{sam}$ representing two sequences, together its matrix $A_{sam}$}
    \label{fig:sequence_graph_2}
\end{figure}
    \end{example}

    This example illustrates an important property of associative graph
    memory: several sequences can be superimposed within a common graph.
    Such superposition is essential for obtaining a compact associative representation,
    but it also introduces ambiguity during retrieval.
    The retrieval algorithms implemented in \myapp address this ambiguity by using contextual information
    and, for ordered retrieval, additional positional information.

    \subsection{Sequence insertion}\label{subsec:sequence-insertion}

    In the software implementation, sequence insertion does not require
    the user to construct the corresponding adjacency matrices explicitly.
    The user specifies the number of symbols and the sequence length when creating an SSAKG
    object and subsequently provides sequences through the package interface.
    The package constructs and updates the underlying graph representation.

    The interface hides the low-level representation
    from the user while retaining direct access to the graph operations required by the retrieval algorithms.
    The treatment of the underlying memory representation
    and its implementation using $Python$ and $C$ is described in Section~\ref{sec:bits_algorithm}.

    \section{Context-based unordered sequence retrieval}\label{sec:unordered_sequence_retrieval}

    The first retrieval operation implemented by SSAKG 1.0 identifies elements associated with a given context.
    The context consists of several elements known to belong to the target sequence.
    The retrieval operation determines the vertices that are jointly connected to the context vertices.

    The algorithm for retrieval of complete sequences is based on the Hadamard product of rows with numbers that the context contains.
    The result of this product is a binary vector.
    This allows us to determine the set of common neighbors for the vertices representing symbols belonging to the context.
    For example, consider the context $C=\{c_1,c_2 \cdots c_n\}$.
    We denote the individual rows of the matrix $A_{sam}$ that correspond to the context symbols
    by $\{r_1, r_2, \cdots, r_n\}$.
    We can write:
    \begin{equation}
    R_{\odot}=\bigodot_{i=1}^{k} r_i
    = r_1 \odot r_2 \odot \cdots \odot r_k.
    \label{eq:hadamard_pattern}
\end{equation}
    where $\odot$ denotes the Hadamard product.
    The ones in the resulting binary vector $R_{\odot}$ indicate the positions of the symbols found in the target sequence.

    \begin{example}
        This example demonstrates how a context consisting $2$ symbols can be used to retrieve the sequences given
        in Example~\ref{example:1}, namely $\{2,6,1\}$ and $\{6,4,5\}$.
        Let context be:
        $C=\{1,6\}$.
        In this case, the binary vector described by equation~\ref{eq:submatrix_asam} is:
        \begin{equation*}
    R_{\odot}=\bigodot_{i=1}^{k} r_i
    = r_1 \odot r_6.
\end{equation*}
        The vector $R_{\odot}$ is obtained by performing the Hadamard product of the rows
        $v_1$ and $v_6$ of the matrix shown in figure~\ref{fig:sequence_graph_2:adjacency}.
        \begin{equation*}
            R_{\odot}=
            \begin{pmatrix}
                0 & 1 & 0 & 0 & 0 & 0
            \end{pmatrix}
        \end{equation*}
        The resulting vector has a non-zero element at position $2$.
        Our set of elements consists of $C=\{1,6\}$ and the element found, $\{2\}$:
        \begin{equation*}
            S = \{2\} \cup C = \{2,1,6\}
        \end{equation*}
    \end{example}
    It should be noted that this memory structure returns the addresses of symbols related to the vertices represented
    by the context.
    However, a minor modification is all that is needed to return the entire set of sequence elements.
    All that is required is to add loops connecting the vertices to one another:
    \begin{equation*}
        A_{sam}' = A_{sam}+I
    \end{equation*}
    alternatively, instead of the matrix $G_s=K_3$, use the matrix of a complete graph with loops.
    It should be noted that modifications of this kind will be used in subsequent algorithms; however,
    for the purpose of code optimisation, the appropriate matrix elements are always used within the programme.


    Algorithm~\ref{alg:unordered_sequence} summarizes the corresponding retrieval procedure.
    Initially, all graph vertices are considered candidates.
    Each context element then eliminates vertices that are not connected to it.
    The remaining indices identify candidate sequence elements.
    The basic representation can also be modified so that the
    context vertices themselves are returned as part of the retrieved set.
    This can be achieved by adding self-connections to the graph representation:
    In the implementation, the corresponding matrix elements
    are modified directly rather than explicitly constructing a new matrix.
    This avoids unnecessary memory operations.
    \begin{algorithm}[H]
        \label{alg:unordered_sequence}
        \scriptsize
        \DontPrintSemicolon
        \caption{Pseudocode of the retrieving unordered sequence elements.}
        \GetUnorderedSequence{$(A_{sam}, context$)}{
            \KwIn{$A_{sam}$}
            \KwIn{context - several elements of the sequence (in random order)}
            \KwOut{all elements of the sequence (in random order)}
            n $\gets$ dimension($A_{sam}$)\;
            \tcp{First, we assume that all possible indices are desired elements of the sequence.}
            unordered\_elements\_row[1..n] $\gets$ $\textbf{1}$\;

            \ForEach{i in context}{
                \tcp{Matrix element $A_{sam}[i]$ can be modified for efficiency}
                context\_row $\gets$ $A_{sam}$[i]\;
                \tcp{Hadamard product of unordered\_elements with context\_row reset unwanted elements}
                unordered\_elements\_row $\gets$ unordered\_elements\_row $\odot$ context\_row\;
            }
            \BlankLine

            unsorted\_elements\_indices $\gets$ indices of unordered\_elements\_row that still contain number different from $\textbf{0}$\;
            \tcp{Finally, we retrieve elements with appropriate indices from the sam matrix.}
            \tcp{The elements have values that are relevant for sorting the sequence.}
            unsorted\_elements $\gets$ $A_{sam}[0,unsorted\_elements\_indices]$\;
            \Return unsorted\_elements

        }
    \end{algorithm}

    \section{Ordered sequence retrieval}
    \label{sec:sam_ordering}

\begin{figure}[H]
    \centering
    \begin{tikzpicture}[scale=0.6]
    \GraphInit[vstyle=Normal]
    \SetUpEdge[style={->}]
    \SetGraphUnit{1}
    \begin{scope}
        \Vertices{circle}{1,2,3}
    \end{scope}

    \Edge(1)(2)
    \Edge(1)(3)
    \Edge(2)(3)
    \end{tikzpicture}
    \caption{Transitive tournament}
    \label{fig:transitive_tournament}
\end{figure}
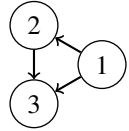

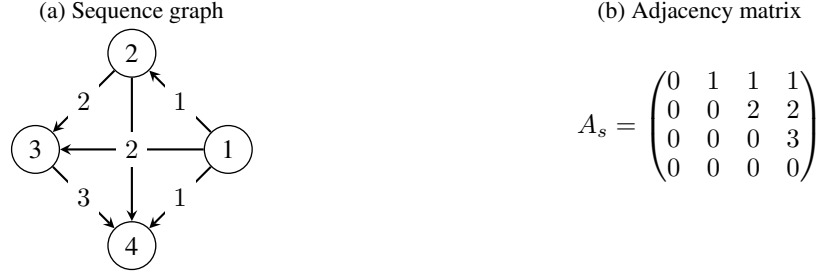
\begin{figure}[H]
        \centering

        \begin{subfigure}[t]{0.45\textwidth}
            \centering
            \caption{Sequence graph}
            \begin{tikzpicture}[scale=0.85,->,>=stealth]
            \GraphInit[vstyle=Normal]
            \SetGraphUnit{1.5}

            \Vertices{circle}{1,2,3,4}

            \Edge[label=$1$](1)(2)
            \Edge[label=$1$](1)(3)
            \Edge[label=$1$](1)(4)

            \Edge[label=$2$](2)(3)
            \Edge[label=$2$](2)(4)

            \Edge[label=$3$](3)(4)
            \end{tikzpicture}
        \end{subfigure}
        \begin{subfigure}[t]{0.45\textwidth}
            \centering
            \caption{Adjacency matrix}
            \[
                A_s=
                \begin{pmatrix}
                    0&1&1&1\\
                    0&0&2&2\\
                    0&0&0&3\\
                    0&0&0&0
                \end{pmatrix}
            \]
        \end{subfigure}
        \caption{A transitive tournament with appropriate weights representing a sequence, with a adjacency matrix.}
        \label{fig:transitive_tournament_weighted}
    \end{figure}

    The unordered representation described above does not preserve the order of sequence elements.
    SSAKG therefore uses a weighted directed when ordered retrieval is required.
    The representation assigns different connection weights according to the positions of elements within the sequence.
    A transitive tournament provides the basic directed structure: for a sequence,
    directed connections are established from the first element to the subsequent
    elements and from the second element to the third.
    No outgoing connection is required from the last element.
    An example of such a graph that corresponds to the order $1\to 2\to 3$.
    is shown in the figure~\ref{fig:transitive_tournament}.
    For retrieval of ordered sequences, the weighted transitive
    tournament shown in Fig~\ref{fig:transitive_tournament_weighted} proved to be more effective.
    Weights are used to filter out edges belonging to other sequences.
    It should be noted that a combination of matrices was used to read the sequence based on the context:
    \begin{equation*}
        A_{sam}' = A_{sam}+A^T_{sam}+I
    \end{equation*}
    With the matrix $A_{sam}$ modified in this way, we can use
    equation~\ref{eq:hadamard_pattern} or the algorithm~\ref{alg:unordered_sequence} to recover the unordered elements of the sequence.
    As noted before, the programme does not add matrices but uses modified matrix elements.

    After receiving the unordered elements of the sequence (associated with the vertices $\{v_1, v_2, \cdots, v_i\}$),
    we proceed to determine their order.
    Using equation~\ref{eq:submatrix_asam}, we retrieve the corresponding matrix $A_{seq}$ from the matrix $A_{sam}$
    and use an appropriate algorithm to determine the order of the elements.
    In the ideal case the matrix $A_{seq}$ is the matrix $A_{s}$ with a different order of rows.
    As additional sequences are stored, however, their connections become superimposed in the associative memory.
    Consequently, the matrix $A_{seq}$ may contain connections originating from other sequences.
    The software therefore requires a node-ordering procedure capable of identifying the ordering corresponding to
    the target sequence despite these additional connections.
    In the programme, this is achieved using the algorithm~\ref{alg:node_ordering}

    \subsection{Node-ordering algorithm}\label{subsec:node-ordering-algorithm}
    Algorithm~\ref{alg:node_ordering} recursively searches for a sequence ordering.
    At each stage, rows containing the expected number of nonzero elements are identified and prioritized according to the selected
    prioritization function.
    The selected row and column are then removed, and the procedure continues with the reduced matrix.
    \begin{algorithm}[H]
        \label{alg:node_ordering}
        \scriptsize
        \DontPrintSemicolon
        \caption{Pseudocode of the Node Ordering.}
        \OrderingAlg{$(A_{seq}, ordering, set\_of\_orderings$)}{
            \KwIn{$A_{seq}$ - submatrix of $A_{sam}$ accociated with set of sequence elemtents}
            \KwIn{ordering - current ordering}
            \KwIn{set\_of\_orderings - set of founded orderings}
            \KwOut{set of founded orderings}
            n $\gets$ dimension($A_{seq}$)\;
            \If{$n>0$}{
                \tcp{In ideal case there is one row}
                non\_zeros\_ $\gets$ find rows contain $(n-1)$ non-zeros elements\;
                \tcp{Prioritize function is different for each algorithm}
                prioritized\_indices $\gets$ \textbf{\CFunction{prioritize}}($A_{seq}$,non\_zeros\_indices)\;
                \ForEach{i in prioritized\_indices}{
                    path.append(i)\;
                    \tcp{The algorithm has found the prioritized rows in the matrix. We can look for the next ones.}
                    $A_{seq}$ $\gets$ remove row i and column i from matrix $A_{seq}$.\;
                    \textbf{node\_ordering}$(A_{seq}, ordering, set\_of\_paths$)\;
                }
            }
            \Else{
                set\_of\_orderings.append(orderings)\;
                \tcp{We are unable to determine which ordering is correct.}
                \tcp{It is desirable for the program to return only one ordering.}
                \tcp{We use the first ordering for all tests.}
                \Return set\_of\_orderings[0] \;
            }

        }
        \;
    \end{algorithm}
    The programme offers a choice of several $priorize$ functions.
    The function presented in algorithm~\ref{alg:prioritize_functions}
    uses the weights assigned to graph edges and prioritizes rows containing the largest relevant weight.
    The prioritization procedure was previously evaluated in~\cite{stoklosa_ssakg};
    the present work incorporates the procedure into the \myapp software implementation.

    This recursive procedure may produce more than one possible ordering when the stored graph
    does not contain sufficient information to uniquely determine the sequence.
    In the implementation used for the experiments, the first ordering returned by the
    procedure is selected.
    This behavior is an implementation choice rather than a
    guarantee that the underlying graph representation always provides a unique reconstruction of the sequence order.

    \begin{algorithm}[H]
        \label{alg:prioritize_functions}
        \scriptsize
        \DontPrintSemicolon
        \caption{Prioritize function}
        \vspace{0.1cm}
        \KwIn{$A_{seq}$}
        \KwIn{non\_zeros\_indices - set of selected indices}
        \KwOut{prioritized indices}
        \tcp{Weighted edges node ordering}
        \Prioritize{$(A_{seq},non\_zeros\_indices)$}{
            \tcp{Determine the prioritised number, in this case the number is equal to the dimension of $A_{seq}$.}
            prioritized\_number $\gets$  dimension($A_{seq}$)\;
            prioritized\_indices $\gets$ find rows with majority of prioritized numbers \;
            \Return prioritized\_indices
        }
        \;
    \end{algorithm}

    \section{Bit-level sequence retrieval}
    \label{sec:bits_algorithm}

    \subsection{Motivation}\label{subsec:motivation}

    The main algorithmic enhancement introduced in \myapp is a bit-level representation of sequence positions.
    Recording each sequence results in the creation of new edges connecting the vertices of the graph $G_{sam}$.
    Context elements belonging to a given sequence may already be associated with other vertices.
    Each context element must be associated with all elements of the sequence.

    As shown in equation~\ref{eq:hadamard_pattern}, the Hadamard product is applied to each element of the context.
    This allows unwanted connections in the matrix $A_{sam}$ to be set to zero for each element of the context.
    In other words, if the context contains $n$ elements, the zeroing operation is performed $n$ times.
    This is especially noticeable in short contexts, where clearing additional elements is less effective.
    This makes it impossible to separate real sequence elements from those associated with other sequences.
    A similar problem was resolved when ordering the elements of the sequence.
    Numerical weights have been added to the relevant edges.
    However, many edges are overwritten when new connections are added.
    Weights appear that make it much more difficult to identify the relevant elements and then sort them.
    One way of significantly increasing memory capacity is
    to introduce weights that employ all the bits in the memory cells.

    \subsection{Bit-based sequence representation}\label{subsec:bit-based-sequence-representation}

    In the \myapp package,  successive positions are represented by distinct powers of two in a single memory cell.
    Each value has only one bit set.
    Consequently, the position information is represented directly by individual bits rather than by a conventional
    numerical weight.
    Figure~\ref{fig:bits_matrix} illustrates the resulting adjacency-matrix representation.
    The package allows sequences of length $64$ to be stored, due to the use of $64$-bit memory cells.

    Memory allocation is described in detail in section~\ref{sec:memory_allocation}.

    \NiceMatrixOptions{
        code-for-first-row = \color{red},
        code-for-first-col = \color{blue},
        code-for-last-row = \color{green},
        code-for-last-col = \color{magenta}
    }
    \begin{figure}[H]
        \begin{equation}
            A_s=
            \begin{pNiceMatrix}[first-row,first-col]
           & & & & & \\
           1 & \mathtt{00001} & \mathtt{00001} & \mathtt{00001}
           & \mathtt{00001} & \mathtt{00001} \\
           2 & \mathtt{00010} & \mathtt{00010} & \mathtt{00010}
           & \mathtt{00010} & \mathtt{00010} \\
           4 & \mathtt{00100} & \mathtt{00100} & \mathtt{00100}
           & \mathtt{00100} & \mathtt{00100} \\
           8 & \mathtt{01000} & \mathtt{01000} & \mathtt{01000}
           & \mathtt{01000} & \mathtt{01000} \\
           16 & \mathtt{10000} & \mathtt{10000} & \mathtt{10000}
           & \mathtt{10000} & \mathtt{10000}
            \end{pNiceMatrix}
            \label{eq:bits_matrix}
        \end{equation}
        \caption{Adjacency matrix used in bits algorithm}
        \label{fig:bits_matrix}
    \end{figure}
    Equation~\ref{eq:submatrix_asam} has been replaced by the corresponding bitwise sum:
    \begin{equation}
        A_{sam}[\{v_1,v_2,\cdots,v_i\}]=A_{sam}[\{v_1,v_2,\cdots,v_i\}]\mid A_s
        \label{eq:submatrix_bits}
    \end{equation}

    We start by calculating the vector $R_{\odot}$ from equation~\ref{eq:hadamard_pattern}.
    As before, we use the algorithm~\ref{alg:unordered_sequence}.
    As in section~\ref{sec:sequences-storage}, we use equation~\ref{eq:submatrix_asam},
    to retrieve the corresponding matrix $A_{seq}$ from the matrix $A_{sam}$.
    However, in this case, the next step will not be to determine the order of the elements.
    When the matrix $A_{sam}$ has a higher density, the vector $R_{\odot}$ contains a number of additional elements;
    the matrix $A_{seq}$ may be considerably larger than the matrix $A_s$.
    The use of bitwise $OR$ allows information represented by different bits to coexist within the same memory cell.
    Thus, connections introduced by different stored sequences can be superimposed without
    requiring a separate numerical value for every combination of connections.

    \subsection{Bit-level filtering}\label{subsec:bit-level-filtering}
    The retrieval process initially follows the procedure described in Section~\ref{sec:unordered_sequence_retrieval}.
    The context identifies candidate sequence elements.
    The purpose of the bit-filtering procedure is to identify the bits
    belonging to the target sequence and eliminate the remaining bits.

    In this case, we start filtering out unwanted connections from the matrix $A_{seq}$.
    Ideally, the matrix $A_{seq}$ is the matrix $A_s$, but the order of the rows in the matrix corresponds to
    the order of the elements in the sequence.
    When a larger number of sequences are stored, in addition to the bits of interest, we obtain a number of extra bits.
    For example, row $i$ of the matrix $A_{seq}$ might look like this:
    \begin{equation*}
        A_{seq}[i]=
        \begin{bNiceMatrix}[first-row,first-col]
          & & & & & & & &\\
          & \mathtt{1010001}
          & \mathtt{1000101}
          & \mathtt{0100100}
          & \mathtt{0100001}
          & \mathtt{0100110}
          & \mathtt{0000011}
          & \mathtt{0101000}
          & \mathtt{0101001}
        \end{bNiceMatrix}
    \end{equation*}
    There will always be an appropriate bit in a row, and the quantity of this bit will always be equal to or greater than the length of the sequence.
    The filtering algorithm initially identifies the bit that appears the most times in a given row.
    We assume that this is the bit we are looking for.
    All other bits are then removed from the row.
    \begin{equation*}
        A_{seq}'[i]=
        \begin{bNiceMatrix}
            \cancel{\mathtt{1}}\mathtt{0}
            \cancel{\mathtt{1}}\mathtt{000}\mathtt{1}
            &
            \cancel{\mathtt{1}}\mathtt{000}
            \cancel{\mathtt{1}}\mathtt{0}\mathtt{1}
            &
            \mathtt{0}\cancel{\mathtt{1}}\mathtt{00}
            \cancel{\mathtt{1}}\mathtt{00}
            &
            \mathtt{0}\cancel{\mathtt{1}}\mathtt{0000}\mathtt{1}
            &
            \mathtt{0}\cancel{\mathtt{1}}\mathtt{00}
            \cancel{\mathtt{1}}\cancel{\mathtt{1}}\mathtt{0}
            &
            \mathtt{00000}\cancel{\mathtt{1}}\mathtt{1}
            &
            \mathtt{0}\cancel{\mathtt{1}}\mathtt{0}
            \cancel{\mathtt{1}}\mathtt{000}
            &
            \mathtt{0}\cancel{\mathtt{1}}\mathtt{0}
            \cancel{\mathtt{1}}\mathtt{00}\mathtt{1}
        \end{bNiceMatrix}
    \end{equation*}
    we obtain:
    \begin{equation*}
        A_{seq}''[i]=
        \begin{bNiceMatrix}
            \mathtt{0000001} &
            \mathtt{0000001} &
            \mathtt{0000000} &
            \mathtt{0000001} &
            \mathtt{0000000} &
            \mathtt{0000001} &
            \mathtt{0000000} &
            \mathtt{0000001}
        \end{bNiceMatrix}
    \end{equation*}

    This bit is treated as the candidate bit corresponding to the current sequence position.
    We remove this particular bit found from the remaining rows of the matrix $A_{seq}$ (it should not be present in the other rows).
    This allows us to immediately remove unwanted edges from the part of the matrix that has not yet been filtered.
    We continue the procedure for all remaining rows of the matrix $A_{seq}$.
    Algorithm~\ref{alg:constraind_sequence} implements this filtering operation:
    \begin{algorithm}[H]
        \label{alg:constraind_sequence}
        \scriptsize
        \DontPrintSemicolon
        \caption{Pseudocode of limiting the number of unordered elements in a sequence.}
        \FilterSequenceMatrix{$(A_{seq}$)}{
            \KwIn{$A_{seq}$ - matrix of sequence (part of sam matrix assiciated with unsorted elements)}
            \KwOut{filtered matrix of sequence ($A_f$)}
            n $\gets$ dimension($A_{seq}$)\;
            \ForEach{i in 1..n}{
                unordered\_elements\_row $\gets$ $A_{seq}[i]$\;
                lead\_bit $\gets$ find most common bit in unordered\_elements\_row\;
                \tcp{Remove other bits from elements row}
                unordered\_elements\_row $\gets$ unordered\_elements\_row \& lead\_bit \;
                \If{$i<n$}{
                    \tcp{Remove lead bit from the rest rows of matrix (rows before index \textbf{i} contains only one unique lead bit)}
                    $A_{seq}[i..n]$ $\gets$ $A_{seq}[i..n]$ $\textbf !$ lead\_bit
                }
            }
            $A_f$ $\gets$ $A_{seq}$ \;
            \tcp{return filtered matrix of sequence}
            \Return $A_f$
        }
    \end{algorithm}
    Once the algorithm has been completed, we obtain the matrix $A_f$ containing elements with the correct bits set.
    The individual rows of the matrix contain different bits.
    We can now proceed to the next stage, which involves removing the unwanted elements from the matrix \textbf{$A_{sam}$}.
    To achieve this, we use an algorithm similar to Algorithm ~\ref{alg:node_ordering}.
    In this case, the argument is not the matrix \textbf{$A_{sam}$},
    but the matrix $A_f$, which is the result of the algorithm~\ref{alg:constraind_sequence}.
    In the rows of $A_f$, only the appropriate bits are set, in agreement with the order of the sequence.
    The design of the algorithm ensures that each row of the matrix $A_f$ contains a different bit.
    The result of executing the algorithm is always a single ordering.
    \begin{algorithm}[H]
        \label{alg:ordered_sequence}
        \scriptsize
        \DontPrintSemicolon
        \caption{Pseudocode of the retrieving sorted sequence elements.}
        \GetOrderedSequence{($A_f$)}{
            \KwIn{$A_f$ - filtered matrix}
            \KwOut{ordered elements of sequence $S$}
            n $\gets$ dimension($A_f$)\;
            \tcp{First, we assume that all possible indices are desired elements of the sequence.}
            unordered\_elements\_row[1..n] $\gets$ $\textbf{1}$\;

            \ForEach{i in n}{
                \tcp{each row of matrix A contains particular bit (eg.5) or zero}
                bit\_row $\gets$ $A_f[i]$\;
                \tcp{Hadamard product of unordered\_elements with bit\_row reset unwanted elements}
                unordered\_elements\_row $\gets$ unordered\_elements\_row $\odot$ bit\_row\;
            }
            \BlankLine

            unsorted\_elements\_indices $\gets$ indices of unordered\_elements\_row that still contain number different from $\textbf{0}$\;
            correct\_order\_indices $\gets$ sort($A_f[0]$) by bit\;

            \BlankLine
            ordered\_elements\_indices $\gets$ unsorted\_elements\_indices[correct\_order\_indices]\;
            \Return $A_{sam}[$ordered\_elements\_indices$]$

        }
    \end{algorithm}

    The bit representation also introduces a direct
    relationship between the supported sequence length and the width of the memory cell.
    The current implementation supports $16$, $32$, and $64-bit$ unsigned integer representations.
    Consequently, the $64-bit$ representation supports sequences of up to $64$ positions.

    \section{Implementation and memory management}
    \label{sec:memory_allocation}

    \subsection{Python-C implementation}\label{subsec:python-c-implementation}

    \myapp is implemented primarily in Python using an object-oriented interface,
    to manage memory efficiently.
    Python provides the user-facing API and manages the high-level operations required to create an associative memory,
    insert sequences, and retrieve stored sequences.

    Performance-critical memory operations are implemented in a $C$ extension.

    The extension uses the memory-management mechanisms
    provided by $NumPy$ and represents the associative-memory matrix as a dynamically
    allocated character array.
    The effective memory-cell width is selected according to the maximum sequence length.

    The matrix $A_{sam}$ is allocated as the simplest $char*$ array.
    The sequence is represented by the matrix $A_s$ which is given by equation~\ref{eq:bits_matrix}.
    This requires selecting a memory cell with the correct number of bits.
    Depending on the length of the sequence, the program selects either a a 16-, 32-, or 64-bit type.
    This corresponds to the types $npy\_uint16$, $npy\_uint32$ and $npy\_uint64$.
    The corresponding addresses of the matrix elements $A_{sam}[i,j]$
    are calculated based on the length of the relevant memory cell.
    This allows the same high-level Python interface to operate with different underlying memory representations.
    All bitwise operations are performed directly on the selected integer type.
    Because the C implementation operates on the underlying memory representation, the package avoids the overhead
    that would result from converting the graph into Python-level objects for individual bit
    operations.
    To make effective use of these commands,
    the code must be compiled in three versions using separate $C$ code paths for the
    supported memory-cell widths.
    An example of the corresponding implementation is shown in Listing~\ref{lst:switch-stride}.

    \begin{lstlisting}[style=cStyle, caption={Extension example}, label={lst:switch-stride}]
        switch (stride) {
            case 2: {
                *(npy_uint16 *)(limited_graph + limited_index * stride) =
                    *(npy_uint16 *)(graph_array + graph_index * stride);
                break;
            }
            case 4: {
                *(npy_uint32 *)(limited_graph + limited_index * stride) =
                    *(npy_uint32 *)(graph_array + graph_index * stride);
                break;
            }
            case 8: {
                *(npy_uint64 *)(limited_graph + limited_index * stride) =
                    *(npy_uint64 *)(graph_array + graph_index * stride);
                break;
            }
            default:
                break;
        }
    \end{lstlisting}

    \subsection{Memory allocation}\label{subsec:memory-allocation}

    The memory representation is selected automatically when the
    associative memory is initialized.
    The package’s programme code itself is object-oriented;
    the appropriate memory cell type is selected automatically.
    The selected width is determined by the sequence length,
    while the underlying graph structure is allocated according to the number of symbols.
    This design allows the package to use smaller memory cells for shorter sequences while supporting longer
    sequences through 64-bit cells.
    The approach also makes the bit-level representation independent of the high-level Python interface.

    The package therefore separates three levels of
    operation:
    \begin{enumerate}[label=(\roman*), leftmargin=*]
        \item{the Python-level representation and user interface}
        \item{the graph-level associative-memory representation}
        \item{the low-level memory operations implemented in C}
    \end{enumerate}

    \subsection{Package usage}\label{subsec:package-usage}

    A user can create and populate an associative memory
    without directly interacting with the $C$ extension or with the underlying memory allocation.
    For example, to allocate memory for $SAM$ and write a
    sequence to it, simply execute the following code:

    \begin{lstlisting}[style=pythonStyle, caption={Usage SSAKG package}, label={lst:ssakg-examlpe}]
        from ssakg.ssakg import SSAKG
        ssakg = SSAKG(
           number_of_symbols=20,
           sequence_length=5,
        )
        ssakg.insert([
            [5, 1, 3, 4, 5],
            [1, 2, 3, 8, 10]
        ])
    \end{lstlisting}
    The example illustrates the intended abstraction level of
    the package: the user specifies the number of symbols and the sequence length
    and then inserts sequences through the $insert()$ method.
    The memory representation and the corresponding low-level operations are handled
    internally by \myapp.

    \section{Experimental evaluation}\label{sec:experimental-evaluation}

    \subsection{Experimental setup}\label{subsec:experimental-setup}

    For the first experiment, sequences were
    stored in associative memory and subsequently retrieved using a randomly
    selected context containing six elements.
    The number of stored sequences was varied for different memory sizes.
    A retrieval error was recorded whenever the complete sequence could not be reconstructed.

    This definition of error measures complete sequence reconstruction rather than partial retrieval accuracy.
    Thus,a sequence is considered correctly reconstructed only when all of its elements
    are recovered in the correct form required by the retrieval procedure.

    The bit based algorithm was evaluated against
    the standard algorithm described in section~\ref{sec:sam_ordering}.
    Further tests of the basic algorithm~\ref{sec:sam_ordering} are described
    in~\cite{stoklosa_ssakg} and on the programme’s website~\cite{Sto2024}.
    The experiments used randomly generated numerical sequences with a uniform symbol distribution.
    Each sequence had a length of $15$, while the number of stored sequences and the number of context
    elements were varied as specified in the tables.
    For the first experiment, sequences selected from 2000 elements were stored in associative
    memory and subsequently retrieved using a randomly selected context.
    The number of stored sequences varied for different memory sizes.
    Then all sequences were read using a random context of length $6$.
    An error was recorded whenever the complete sequence could not be reconstructed.
    A sequence is considered correctly reconstructed only when all of its elements
    are recovered in the correct form required by the retrieval procedure.

    \subsection{Effect of the number of stored sequences}\label{subsec:effect-of-the-number-of-stored-sequences}

    Table~\ref{tab:algorithm_comparison_symbols} compares the retrieval error obtained with
    the standard ssakg's algorithms described in~\cite{stoklosa_ssakg,raif2026}, and bit-based
    algorithms as the number of stored sequences increases.
    The results show a substantial difference between the two approaches.
    For $2,000$ possible symbols, the standard algorithm has a retrieval error of 0.0\% with $1,000$ stored sequences,
    increasing to 100\% with $11,000$ stored sequences.
    In contrast, the bit-based algorithm has zero error for all tested conditions except the largest tested load, where the
    error remains only 0.6\%.

    These results indicate that the bit-based representation is substantially more robust to the accumulation of
    additional sequence connections in the associative memory.
    The difference becomes increasingly pronounced as the number of stored sequences increases.
    The result is consistent with the intended purpose of the bit representation:
    individual sequence positions can remain distinguishable even when connections
    from many sequences are superimposed.

    \begin{table}[H]
        \centering
        \caption{Scene recognition error in \% for various dataset sizes for sequence length 15 and context length 6.}
        \label{tab:algorithm_comparison_symbols}

        \begin{NiceTabular}{llrrrrrrrrrrr}[cell-space-limits=2pt]
                \toprule
                &  & 1000 & 2000 & 3000 & 4000 & 5000 & 6000 & 8000 & 10000 & 11000 & 13000 & 15000\\
                No. symbols & Algorithm
                & & & & & & & & & & \\
                \midrule

                \Block{2-1}{2000}
                & standard
                & 0.0 & 0.4 & 2.5 & 11.9 & 30.1
                & 57.5 & 94.5 & 97.9 & 100.0 & 100.0 & 100.0 \\

                & bits
                & 0.0 & 0.0 & 0.0 & 0.0 & 0.0
                & 0.0 & 0.0 & 0.0 & 0.0 & 0.6 & 12.5 \\

                \bottomrule

                \CodeAfter
                \SubMatrix{\lbrace}{3-2}{4-2}{.}

        \end{NiceTabular}
    \end{table}
    \

    \subsection{Effect of context length}\label{subsec:effect-of-context-length}

    A second experiment investigated the effect of context length on sequence reconstruction.
    The number of possible symbols was fixed at $2,000$, the sequence length was $15$, and the context length
    was varied from three to seven elements.
    The results are presented in Table~\ref{tab:algorithm_comparison}.
    The conventional algorithm is strongly affected by the context length.
    For $1,000$ stored sequences, its error decreases from 29.3\% with a context of three
    elements to zero with a context of six or seven elements.
    As the number of stored sequences increases, the degradation becomes substantially stronger.
    For example, with $5,000$ stored sequences, the conventional algorithm has errors
    of 100\%, 99.2\%, 74.8\%, 30.7\%, and 8.5\% for context lengths from three to seven, respectively.
    The bit-based algorithm performs substantially better over the same range of conditions.
    Its error remains zero in the reported tests for the tested context lengths and loads, except for the
    shortest context at the largest memory load, where the reported error is 1.6\%.
    These results demonstrate that the bit-based representation is particularly
    beneficial when contextual information is limited.
    This is important for associative sequence retrieval because short contexts provide
    less information for distinguishing the target sequence from other sequences stored in the same graph.

    \begin{table}[H]
        \centering
        \caption{Scene recognition error in \% for various dataset and context sizes for sequence length 15. No symbols 2000.}
        \label{tab:algorithm_comparison}

        \begin{NiceTabular}{llrrrrr}[cell-space-limits=2pt]
                \toprule
                &  & 3 & 4 & 5 & 6 & 7 \\
                Stored sequences & Algorithm &  &  &  &  &  \\
                \midrule

                \Block{2-1}{1000}
                & standard & 29.3 & 1.8 & 0.2 & 0.0 & 0.0 \\
                & bits     & 0.0  & 0.0 & 0.0 & 0.0 & 0.0 \\

                \Block{2-1}{2000}
                & standard & 84.3 & 23.8 & 2.8 & 0.3 & 0.0 \\
                & bits     & 0.10  & 0.0  & 0.0 & 0.0 & 0.0 \\

                \Block{2-1}{3000}
                & standard & 99.2 & 61.5 & 16.4 & 3.0 & 0.5 \\
                & bits     & 0.3  & 0.0  & 0.0  & 0.0 & 0.0 \\

                \Block{2-1}{4000}
                & standard & 99.9 & 91.1 & 42.7 & 10.8 & 2.5 \\
                & bits     & 0.8  & 0.0  & 0.0  & 0.0  & 0.0 \\

                \Block{2-1}{5000}
                & standard & 100 & 99.2 & 74.8 & 30.7 & 8.5 \\
                & bits     & 1.6  & 0.0  & 0.0  & 0.0  & 0.0 \\

                \bottomrule

                \CodeAfter
                \SubMatrix{\lbrace}{3-2}{4-2}{.}
                \SubMatrix{\lbrace}{5-2}{6-2}{.}
                \SubMatrix{\lbrace}{7-2}{8-2}{.}
                \SubMatrix{\lbrace}{9-2}{10-2}{.}
                \SubMatrix{\lbrace}{11-2}{12-2}{.}

        \end{NiceTabular}
    \end{table}

    \subsection{Interpretation of the results}\label{subsec:interpretation-of-the-results}

    The experiments demonstrate two complementary properties of the bit-based implementation.
    First, its retrieval accuracy degrades much more slowly than that of the conventional implementation
    as additional sequences are stored.
    Second, it maintains substantially better retrieval accuracy when the
    context contains relatively few elements.

    The results support the use of bit-level memory cells as an implementation strategy
    for reducing interference between superimposed sequence representations.
    They do not, however, establish a general theoretical capacity limit or a computational
    speedup relative to the conventional implementation.
    Such conclusions would require additional experiments measuring memory consumption
    and execution time under controlled conditions.

    \section{Discussion}\label{sec:discussion}

    \subsection{Software contribution}\label{subsec:software-contribution}

    The primary contribution of \myapp is the transformation of the previously
    developed SSAKG model into a reusable software package.
    The package provides a $Python$ interface for constructing
    associative sequence memory, inserting sequences, and performing context-based
    retrieval, while performance-critical memory operations are implemented in $C$.

    This separation between the high-level
    interface and low-level memory operations allows users to work with the
    associative-memory model without managing the underlying memory representation directly.
    The package also provides the bit-level retrieval algorithms introduced in this version.

    The experimental results indicate that this implementation provides a practical
    mechanism for reducing retrieval errors caused by the superposition
    of connections from multiple stored sequences.

    \subsection{Limitations}\label{subsec:limitations}

    The bit-level representation has an inherent limitation: the maximum sequence
    length is related to the number of available bits in the memory cell.
    The current implementation supports 64-bit cells and therefore sequences of up to $64$ positions.

    A second limitation follows from the filtering strategy.
    The algorithm assumes that the bit associated with the
    target sequence position can be identified from its frequency within the corresponding row.
    Under sufficiently high graph density or particular patterns
    of interference, this assumption may no longer hold.

    \textbf{The present experimental evaluation is also limited to randomly generated numerical sequences}.
    Although the manuscript refers to evaluations using $NLTK$ sentence
    sequences and $mRNA$ sequences in the abstract/introduction, corresponding
    experimental results are not included in the current Results section.
    These examples demonstrate that the same memory mechanism can be applied to substantially
    different classes of symbolic sequences.

    Finally, the current evaluation focuses on reconstruction accuracy.
    It does not yet provide systematic measurements of execution time,
    memory consumption, scalability with respect to
    graph size, or comparison with alternative sequence-memory implementations.

    \subsection{Reproducibility and extensibility}\label{subsec:reproducibility-and-extensibility}

    \myapp is distributed as open-source software under the \texttt{Apache 2.0}.
    The source code and examples are available through $GitHub$, while
    the ssakg package is distributed through $PyPI$.

    The combination of an object-oriented $Python$ interface, a low-level $C$ extension,
    and examples for sequence insertion and retrieval is intended to support reuse and further development.
    The software architecture also allows alternative retrieval procedures to be
    incorporated without requiring users to change the high-level representation of
    the associative memory.

    For a software-oriented publication,
    reproducibility depends not only on providing source code but also on providing
    the exact software version, dependencies, experimental scripts, and
    data-generation procedures used to obtain the reported results.
    These items should therefore accompany the final release associated with this article.

    \subsection{Future development}\label{subsec:future-development}
    Future development can address several limitations of the current implementation.
    These include extending the supported sequence length, improving retrieval under higher memory loads,
    evaluating larger and more diverse datasets,
    and systematically benchmarking execution time and memory consumption.

    The open-source architecture also allows alternative graph representations and retrieval
    procedures to be implemented and compared within the same software framework.

    \section{Conclusions}\label{sec:conclusions}

    \myapp provides an open-source implementation of Structural Associative Sequence Knowledge Graph
    memory for storing and retrieving sequences from contextual information.
    The package implements the previously established SSAKG model and provides a $Python$
    interface combined with a $C$ extension for efficient manipulation of the
    underlying graph memory.
    A principal enhancement in version $2.0$ is the bit-level representation of sequence positions.
    The new memory allocation method enabled the design and implementation of significantly faster and more efficient
    algorithms, resulting in a several-fold increase in associative memory capacity and improved accuracy.
    The new memory allocation method enabled the design and implementation of significantly faster and more efficient algorithms,
    resulting in a several-fold increase in associative memory capacity and improved accuracy.
    The experimental results show substantially lower
    reconstruction error for the bit-based algorithm than for the conventional
    implementation, particularly as the number of stored sequences increases and
    the available context becomes shorter.

    The package is distributed as open-source software under the \texttt{Apache 2.0} license,
    with source code and examples available through GitHub and the \texttt{Python} package available through $PyPI$.
    \myapp is intended not only as an implementation of the existing model but also as
    a reusable software framework for further investigation of structural
    associative sequence memory.

    \bibliographystyle{unsrt}
    \bibliography{bibliography-bibtex}
\end{document}